\documentclass[letterpaper,journal]{IEEEtran}
\ifdefined\pdfminorversion\pdfminorversion=7\fi

\usepackage{amsmath,amsfonts,amssymb,bm}
\usepackage{algorithmic}
\usepackage{algorithm}
\usepackage{array}
\usepackage[caption=false,font=normalsize,labelfont=sf,textfont=sf]{subfig}
\usepackage{textcomp}
\usepackage{stfloats}
\usepackage{url}
\usepackage{verbatim}
\usepackage{graphicx}
\usepackage{cite}
\usepackage{booktabs}
\usepackage{multirow}
\usepackage{xcolor}
\usepackage{enumitem}
\usepackage[hidelinks]{hyperref}

\setlist[itemize]{leftmargin=1.45em,itemsep=0.35em,topsep=0.45em,parsep=0pt}
\title{FFAI: Flow Matching with Feedback Training and Adaptive Initialization for CT Reconstruction}

\author{Jigang~Duan$^{\dagger}$,
Heran~Wang$^{\dagger}$,
Genwei~Ma,
and Xing~Zhao$^{*}$%
\thanks{$^{\dagger}$Jigang Duan and Heran Wang contributed equally to this work.}%
\thanks{$^{*}$Corresponding author: Xing Zhao
(e-mail: zhaoxing\_1999@126.com).}%
\thanks{Jigang Duan, Heran Wang, and Xing Zhao are with the
School of Mathematical Sciences, Capital Normal University,
Beijing 100048, China
(e-mails: hiduanjigang@163.com (Jigang Duan);
2250501041@cnu.edu.cn (Heran Wang)).}%
\thanks{Genwei Ma is with the National Center for Applied Mathematics Beijing,
Capital Normal University, and the Academy for Multidisciplinary Studies,
Capital Normal University, Beijing 100048, China
(e-mail: magenwei@126.com).}}

\hypersetup{
  pdftitle={FFAI: Flow Matching with Feedback Training and Adaptive Initialization for CT Reconstruction},
  pdfauthor={Jigang Duan, Heran Wang, Genwei Ma, Xing Zhao},
  pdfkeywords={computed tomography, conditional flow matching, degradation-adaptive initialization, feedback-aware training, image reconstruction}
}

\begin{document}
\maketitle

\begin{abstract}
Although low-dose, sparse-view, and limited-angle CT involve distinct degradation mechanisms, all three pose reconstruction challenges due to limited measurement information and can therefore be addressed within a unified framework. In iterative image restoration with flow matching, prediction errors may cause inference states to deviate from the prescribed training path, while a fixed initialization may be unsuitable for inputs with varying degradation severity. We propose a unified conditional flow matching method that conditions on the initial reconstruction and combines feedback-aware training with degradation-adaptive initialization. During training, a teacher network simulates inference updates to generate shifted states, from which a student network learns to recover high-quality reference images, improving its ability to handle prediction errors. During inference, the starting time is selected based on the noise level or extent of missing projection data, and the noise added to the initial state is adjusted accordingly to adapt the starting point to the input. Experiments on public benchmark datasets show that the proposed method achieves better quantitative results than the compared methods across all three reconstruction tasks, effectively suppressing noise and artifacts while preserving anatomical structures and details.
\end{abstract}


\section{Introduction}

\IEEEPARstart{C}{omputed} tomography (CT) provides cross-sectional images of internal anatomical structures and plays an important role in disease screening and clinical diagnosis. Maintaining image quality while reducing radiation dose remains a central goal in medical CT\cite{smithbindman2025risk}. Radiation dose is typically reduced by decreasing either the incident photon count per view or the number of projection views. The former yields low-dose CT, whereas the latter can be achieved by increasing the angular sampling interval or reducing the scan angular range, corresponding to sparse-view and limited-angle CT, respectively\cite{ge2026tamp}. Despite their distinct degradation mechanisms, all three conditions provide limited measurement information, weakening the constraints that measured projections impose on the reconstructed image. We collectively refer to these conditions as degraded acquisition and investigate a unified reconstruction method.

\subsection{Trends in Models and Algorithms}

Model-based iterative reconstruction (MBIR) combines data fidelity and image priors to constrain reconstruction. For low-dose CT, reconstruction can use a likelihood derived from measurement statistics or a penalized weighted least-squares formulation\cite{elbakri2002statistical,wang2006pwlsct}. For sparse-view and limited-angle CT, the projection operator is defined by the acquired angles. Priors such as total variation, prior-image constraints, and learned sparsifying transforms further compensate for insufficient measurements\cite{sidky2008tv,chen2008piccs,yang2021mars,chen2023mcst}. These methods explicitly combine measurements with prior information, but their performance remains dependent on the expressiveness of the prior and the choice of regularization parameters. Preserving details while removing noise and artifacts remains challenging.

Deep learning methods learn richer image features from data. Direct mapping methods recover high-quality images in the image domain, projection domain, or both\cite{shen2023mlfiosc,yang2023sist,li2022ddptransformer}. Noise2Noise can also learn restoration mappings from independent noisy observations of the same signal, reducing reliance on clean targets\cite{lehtinen2018noise2noise}. However, without explicit measurement constraints during inference, conditioning on a degraded input alone does not guarantee consistency with the measured projections. Model-driven networks integrate data-fidelity updates with learned modules, allowing measurements to guide multiple stages of restoration\cite{ye2021super,wu2021drone,hu2022dior,pan2024iron}. To accommodate different acquisition conditions, ProCT uses view prompts for sparse-view and limited-angle reconstruction, while PromptCT incorporates sampling prompts into an unrolled network to handle multiple sparse-view configurations\cite{ma2026proct,shi2026promptct}. TAMP addresses all three degradation conditions through large-scale simulation-based pretraining, dual-domain supervision, and parameter-efficient fine-tuning\cite{ge2026tamp}. These studies provide a foundation for unified reconstruction under different acquisition conditions, but adaptation across conditions and error control during iterative restoration remain distinct challenges.

Generative models learn the distribution of high-quality images to provide priors for structural recovery from insufficient measurements. Generative adversarial networks have been used to enhance detail and texture in low-dose CT\cite{zhang2021clear,fu2023aigan}, while diffusion, score-based generative, and consistency models have been applied to low-dose, sparse-view, and limited-angle reconstruction\cite{liao2025ipdm,hein2025pfcm,xu2024sword,yang2025ctsdm,zhang2024wism,chen2026founddiff}. However, an image consistent with the training distribution may not accurately represent the imaged object, and a mismatch between the prior and the object can produce spurious structures\cite{bhadra2021hallucinations}. In multistep generation, each prediction also affects subsequent inputs, allowing prediction errors to propagate\cite{gao2024corediff}. Applying generative priors to reconstruction therefore requires attention to the network's ability to handle the states encountered during inference.

Previous studies have strengthened measurement guidance by jointly optimizing generative priors and data-fidelity terms, incorporating measurement constraints into generative updates, or integrating iterative reconstruction with generative sampling\cite{he2025vss,wu2024mogm,han2025psdm,chung2023dps,li2025tlir}. FORCE combines a PFGM++ prior with task-specific measurement correction for low-dose CT, sparse-view CT, and metal artifact correction\cite{xia2026force}. These methods demonstrate that generative priors can accommodate different observation models. However, measurement correction also changes the state passed to the network at the next step. A network trained only on prescribed noisy or interpolated states may not fully accommodate these changes.

To address state changes during iterative restoration, CoreDiff re-degrades network predictions for training, enabling the restoration network to handle subsequent inputs derived from its own predictions\cite{gao2024corediff}. CDDM trains an additional discrepancy mitigation model within a cascaded diffusion framework to handle states after data-consistency correction\cite{chen2025cddm}. These studies indicate that, in addition to incorporating measurement constraints at inference, training should cover the states produced by the actual updates. Flow matching (FM) learns a velocity field connecting source and target distributions, providing a convenient formulation for this joint design\cite{lipman2023flow}. Along a linear conditional path, the current state and predicted velocity directly yield an estimate of the target endpoint. FLOWER uses this formulation to solve inverse problems by combining endpoint prediction, measurement correction, and state advancement\cite{pourya2026flower}. Here, we further investigate state shifts caused by sequential predictions and the suitability of a fixed initialization for inputs with varying degradation severity. Feedback-aware training and degradation-guided initialization connect the requirements of inference with the existing FM training path.

\subsection{Motivation}

\begin{figure}[!t]
    \centering
    \includegraphics[width=\columnwidth]{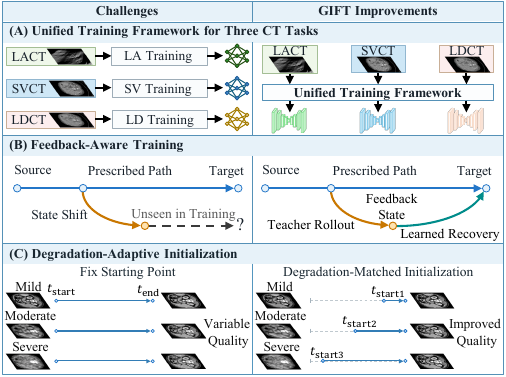}
    \caption{Motivation and key mechanisms of FFAI.
(A) The three CT tasks share a unified training procedure, with parameters learned separately for each task.
(B) Feedback-based error correction: shifted states generated by teacher-simulated inference are incorporated into training, with high-quality reference images supervising restoration to improve the network's ability to handle prediction errors.
(C) Degradation adaptation: the starting time and initial state are jointly determined by the input degradation severity to accommodate different restoration needs.}
    \label{fig:motivation}
\end{figure}

Although the three CT tasks involve different degradation mechanisms, their initial reconstructions all retain some structural information about the imaged object. Based on this shared property, we develop FFAI, a unified conditional FM method that uses the initial reconstruction and acquisition conditions to guide restoration. As shown in Fig.~\ref{fig:motivation}(A), the three tasks share the same training objective and inference rules, with parameters learned separately for each task. Feedback-based error correction during training mitigates the effects of prediction errors, while degradation adaptation during inference selects an appropriate starting point for each input.

Training should equip the network to correct prediction errors. Standard FM training uses states sampled from a prescribed path, whereas each inference input depends on the preceding prediction. Prediction errors may shift subsequent inputs away from the training-state distribution, making them difficult for the network to handle and allowing errors to propagate, as illustrated on the left of Fig.~\ref{fig:motivation}(B). To address this issue, a teacher network simulates inference updates, and the resulting shifted states serve as additional training inputs. Using high-quality reference images as targets, the restoration network learns to continue restoration from these states, as shown on the right of Fig.~\ref{fig:motivation}(B). Training on both prescribed path states and teacher-generated feedback states teaches the network the restoration process as well as how to handle deviations encountered along the way.

Inference should adapt its starting point to the input degradation severity. Initial reconstructions with different degradation levels retain different amounts of information. A fixed starting configuration may therefore be unsuitable for some inputs and lead to substantial variation in restoration quality, as shown on the left of Fig.~\ref{fig:motivation}(C). We quantify degradation severity using the noise level or sampling conditions and use it to select a starting time on the existing FM training path. The initial reconstruction is then used to construct an inference input that approximately matches the training state at that time. As illustrated on the right of Fig.~\ref{fig:motivation}(C), this degradation-adaptive initialization selects a corresponding starting time and image state for each input, adapting restoration to its degradation severity.

\subsection{Our Contributions}

The main contributions are as follows:
\begin{itemize}
\item We develop FFAI, a unified conditional FM method for low-dose, sparse-view, and limited-angle CT, in which the initial reconstruction and acquisition conditions jointly guide restoration.
\item We design feedback-aware training guided by an exponential moving average (EMA) teacher, enabling the network to recover target images from shifted states produced by sequential predictions.
\item We introduce a quantitative, degradation-guided initialization that jointly determines the starting time and initial state on the existing FM training path to accommodate reconstruction at different degradation levels.
\end{itemize}

\begin{figure*}[!t]
    \centering
    \includegraphics[width=\textwidth]{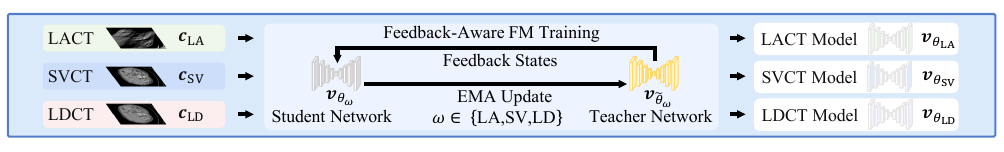}
    \caption{Unified training framework for the three tasks. Each task uses the initial reconstruction as a condition and shares the same flow matching and feedback training mechanisms, while network parameters are learned independently. After training, the student model for each task is retained for inference.}
    \label{fig:task_framework}
\end{figure*}

\section{Methodology}
\label{sec:method}

\subsection{Conditional Flow Matching for CT Reconstruction}
\label{subsec:conditional_fm}

FFAI progressively restores initial CT reconstructions obtained under degraded acquisition conditions using a conditional velocity field. As shown in Fig.~\ref{fig:task_framework}, low-dose, sparse-view, and limited-angle CT share the same training objective and inference rules, with parameters learned separately for each task. The model first learns the restoration velocity field through standard FM and then learns to handle prediction errors using feedback states. During inference, the starting point is determined by the input degradation severity, and the image is progressively restored along the learned velocity field.

Let $\omega\in\{\mathrm{LD},\mathrm{SV},\mathrm{LA}\}$ denote the task, $\boldsymbol{y}_{\omega}$ the projection measurements, and $\boldsymbol{x}^{\star}$ the paired high-quality reference image. An initial reconstruction is obtained using a direct reconstruction operator $\mathcal{B}_{\omega}$, and a Gaussian perturbation is added to construct the source endpoint for training:
\begin{equation}
\begin{aligned}
\boldsymbol{c}_{\omega}
&=\mathcal{B}_{\omega}(\boldsymbol{y}_{\omega}),\\
\boldsymbol{z}_{\omega}
&=\boldsymbol{c}_{\omega}+\gamma_{\omega}\boldsymbol{\epsilon},
\qquad \boldsymbol{\epsilon}\sim\mathcal{N}(\boldsymbol{0},\boldsymbol{I}),
\end{aligned}
\label{eq:method_source}
\end{equation}
where $\gamma_{\omega}\geq0$ controls the perturbation amplitude at the source endpoint. All images are normalized consistently. The linear path connecting the reference image and the source endpoint is
\begin{equation}
\boldsymbol{x}_{t}
=(1-t)\boldsymbol{x}^{\star}+t\boldsymbol{z}_{\omega},
\qquad t\in[0,1].
\label{eq:method_path}
\end{equation}
The path corresponds to the source endpoint at $t=1$ and the reference image at $t=0$; reconstruction therefore proceeds backward in time.

We describe a single task below and abbreviate its parameters $\theta_{\omega}$ as $\theta$. The target velocity along the path is $\boldsymbol{u}_{\omega}=\boldsymbol{z}_{\omega}-\boldsymbol{x}^{\star}$. The velocity network takes the current state $\boldsymbol{x}_t$, time $t$, and initial reconstruction $\boldsymbol{c}_{\omega}$ as inputs and is trained by minimizing
\begin{equation}
\mathcal{L}_{\mathrm{FM}}
=\mathbb{E}\!\left[
\left\|
\boldsymbol{v}_{\theta}(\boldsymbol{x}_{t},t\mid\boldsymbol{c}_{\omega})
-\boldsymbol{u}_{\omega}
\right\|_2^2
\right].
\label{eq:method_fm_loss}
\end{equation}
The expectation is taken over training samples, random perturbations, and sampled times. The image condition $\boldsymbol{c}_{\omega}$ remains fixed throughout restoration, providing object-specific information for velocity prediction at each step.

\subsection{Feedback-Aware FM Training}
\label{subsec:feedback_training}

\begin{figure*}[!t]
    \centering
    \includegraphics[width=\textwidth]{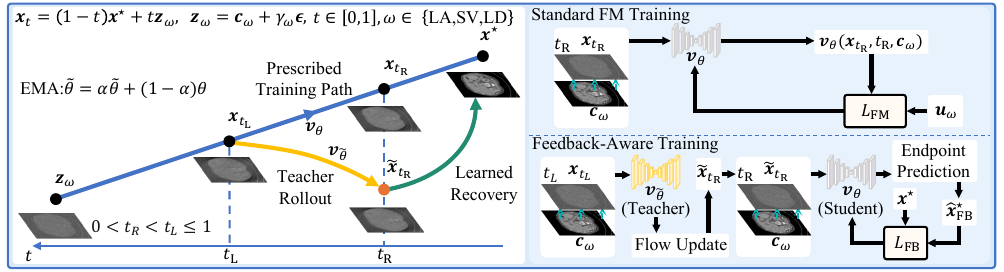}
    \caption{Feedback-aware FM training. The standard branch learns the target velocity at the prescribed path state $\boldsymbol{x}_{t_R}$. In the feedback branch, the EMA teacher performs a single update from $\boldsymbol{x}_{t_L}$ to generate a stop-gradient feedback state $\widetilde{\boldsymbol{x}}_{t_R}$, and the student's endpoint prediction is supervised by the reference image. $t_L$ and $t_R$ are the nearest inference nodes on either side of the randomly sampled time $t$.}
    \label{fig:feedback_training}
\end{figure*}

Standard FM trains on states sampled from a prescribed path, whereas inference states are generated by successive network predictions. To accommodate state shifts caused by prediction errors, FFAI uses two-stage training. First, an initial network is trained by minimizing only $\mathcal{L}_{\mathrm{FM}}$. This network then serves as the student, and an EMA teacher is initialized with the same parameters. Feedback training is subsequently introduced while retaining standard FM supervision. Fig.~\ref{fig:feedback_training} shows the two training branches in the second stage.

To match the time interval of the feedback update to that used during inference, we define the inference grid as $\mathcal{S}=\{s_j=j/(N-1)\}_{j=0}^{N-1}$, where $N$ is the number of nodes. During training, $t$ is randomly sampled from a denser time grid, and the two adjacent inference nodes bracketing $t$ are selected, with $t_L$ denoting the larger time and $t_R$ the smaller time, such that $0<t_R<t<t_L\leq1$. The states $\boldsymbol{x}_{t_R}$ and $\boldsymbol{x}_{t_L}$ are constructed using \eqref{eq:method_path} with the same training sample and random perturbation. The standard branch evaluates \eqref{eq:method_fm_loss} at $t_R$.

In the feedback branch, the teacher performs one Euler update from $\boldsymbol{x}_{t_L}$:
\begin{equation}
\widetilde{\boldsymbol{x}}_{t_R}
=\operatorname{sg}\!\bigl[\boldsymbol{x}_{t_L}+(t_R-t_L)\
\boldsymbol{v}_{\bar{\theta}}
(\boldsymbol{x}_{t_L},t_L\mid\boldsymbol{c}_{\omega})\bigr],
\label{eq:method_feedback_state}
\end{equation}
where $\bar{\theta}$ denotes the teacher parameters and $\operatorname{sg}[\cdot]$ denotes the stop-gradient operation. The deviation of this state from the prescribed path satisfies
\begin{equation}
\widetilde{\boldsymbol{x}}_{t_R}-\boldsymbol{x}_{t_R}
=(t_R-t_L)\bigl[
\boldsymbol{v}_{\bar{\theta}}
(\boldsymbol{x}_{t_L},t_L\mid\boldsymbol{c}_{\omega})
-\boldsymbol{u}_{\omega}\bigr].
\label{eq:method_feedback_deviation}
\end{equation}
Thus, the feedback state in Fig.~\ref{fig:feedback_training} reflects the teacher's actual prediction error, exposing the student during training to shifted inputs generated by a single inference step.

The original path velocity label no longer applies to a shifted state. Using the endpoint relation of the linear path, we convert the student's predicted velocity into a target image estimate and supervise it directly with the reference image:
\begin{equation}
\begin{aligned}
\widehat{\boldsymbol{x}}_{\mathrm{FB}}
&=\widetilde{\boldsymbol{x}}_{t_R}
-t_R\boldsymbol{v}_{\theta}
(\widetilde{\boldsymbol{x}}_{t_R},t_R\mid\boldsymbol{c}_{\omega}),\\
\mathcal{L}_{\mathrm{FB}}
&=\mathbb{E}\!\left[
\left\|\widehat{\boldsymbol{x}}_{\mathrm{FB}}
-\boldsymbol{x}^{\star}\right\|_2^2\right].
\end{aligned}
\label{eq:method_feedback_loss}
\end{equation}
The teacher only generates the inputs; the supervision target remains $\boldsymbol{x}^{\star}$. Gradients are stopped at the feedback input, so this loss updates only the student parameters.

The joint objective for the second stage is
\begin{equation}
\mathcal{L}
=\mathcal{L}_{\mathrm{FM}}
+\lambda_{\mathrm{FB}}\mathcal{L}_{\mathrm{FB}},
\label{eq:method_joint_loss}
\end{equation}
where $\lambda_{\mathrm{FB}}>0$ balances path velocity supervision and feedback endpoint supervision. After each student update, the teacher parameters are updated as
\begin{equation}
\bar{\theta}\leftarrow
\alpha\bar{\theta}+(1-\alpha)\theta,
\qquad \alpha\in[0,1).
\label{eq:method_ema}
\end{equation}
After training, the student parameters $\theta_{\omega}$ are retained as the task model. Algorithm~\ref{alg:FFAI_training} summarizes the two-stage training procedure.

\begin{algorithm}[t]
\caption{Two-Stage Feedback-Aware FM Training}
\label{alg:FFAI_training}
\begin{algorithmic}[1]
\REQUIRE Task $\omega$, paired training data, inference grid $\mathcal{S}$, training iterations $T_{\mathrm{pre}}$ and $T_{\mathrm{fb}}$
\ENSURE Task model parameters $\theta_{\omega}$
\STATE Initialize student parameters $\theta$
\STATE \textbf{Stage 1: Standard FM Pretraining}
\FOR{$n=1$ to $T_{\mathrm{pre}}$}
    \STATE Sample a training pair, perturbation $\boldsymbol{\epsilon}$, and time $t$
    \STATE Construct $\boldsymbol{c}_{\omega}$, $\boldsymbol{z}_{\omega}$, and $\boldsymbol{x}_t$ using \eqref{eq:method_source}--\eqref{eq:method_path}
    \STATE Update $\theta$ by minimizing $\mathcal{L}_{\mathrm{FM}}$
\ENDFOR
\STATE \textbf{Stage 2: Feedback-Aware Training}
\STATE Initialize the teacher $\bar{\theta}\leftarrow\theta$
\FOR{$n=1$ to $T_{\mathrm{fb}}$}
    \STATE Sample a training pair, perturbation $\boldsymbol{\epsilon}$, and time $t$
    \STATE Select the nearest inference nodes on either side of $t$, satisfying $0<t_R<t<t_L\leq1$
    \STATE Construct $\boldsymbol{x}_{t_L}$ and $\boldsymbol{x}_{t_R}$ on the same path
    \STATE Compute $\mathcal{L}_{\mathrm{FM}}$ for the student at $\boldsymbol{x}_{t_R}$
    \STATE Generate the feedback state using the teacher via \eqref{eq:method_feedback_state}
    \STATE Compute $\mathcal{L}_{\mathrm{FB}}$ for the student using \eqref{eq:method_feedback_loss}
    \STATE Update $\theta$ by minimizing \eqref{eq:method_joint_loss}
    \STATE Update $\bar{\theta}$ using \eqref{eq:method_ema}
\ENDFOR
\STATE $\theta_{\omega}\leftarrow\theta$
\end{algorithmic}
\end{algorithm}

\subsection{Degradation-Adaptive Inference}
\label{subsec:adaptive_inference}

\begin{figure*}[!t]
    \centering
    \includegraphics[width=\textwidth]{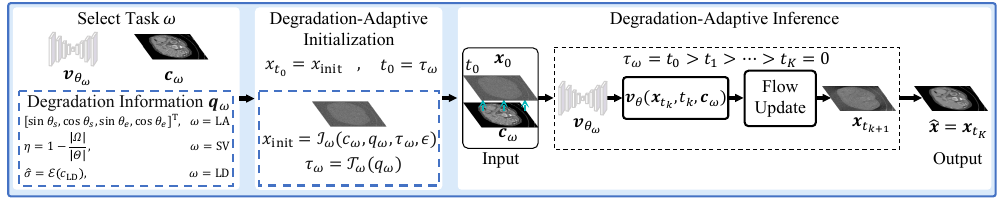}
    \caption{Degradation-adaptive inference. The starting time $\tau_{\omega}$ is selected from the degradation descriptor $\boldsymbol{q}_{\omega}$, and the corresponding initial state is constructed. Conditioned on the fixed image $\boldsymbol{c}_{\omega}$, the task-specific student network performs Euler updates along a decreasing time grid to obtain the final reconstruction.}
    \label{fig:adaptive_inference}
\end{figure*}

The information retained in the initial reconstruction varies with degradation severity. FFAI adjusts the starting point accordingly: milder degradation corresponds to a smaller starting time, and more severe degradation to a larger starting time. As shown in Fig.~\ref{fig:adaptive_inference}, the degradation descriptor $\boldsymbol{q}_{\omega}$ determines the restoration interval through the time mapping $\tau_{\omega}=\mathcal{T}_{\omega}(\boldsymbol{q}_{\omega})$, and the selected time is then used to construct the initial image.

The initial image is constructed to match the noise scale of the training path. From \eqref{eq:method_path}, the state at time $\tau_{\omega}$ is $(1-\tau_{\omega})\boldsymbol{x}^{\star}+\tau_{\omega}\boldsymbol{c}_{\omega}+\tau_{\omega}\gamma_{\omega}\boldsymbol{\epsilon}$. Since the reference image is unavailable during inference, we approximate $\boldsymbol{x}^{\star}$ with $\boldsymbol{c}_{\omega}$, yielding
\begin{equation}
\begin{aligned}
\boldsymbol{x}_{\mathrm{init}}
&=\mathcal{I}_{\omega}
(\boldsymbol{c}_{\omega},\boldsymbol{q}_{\omega},
\tau_{\omega},\boldsymbol{\epsilon})\\
&=\boldsymbol{c}_{\omega}
+\tau_{\omega}\gamma_{\omega}\boldsymbol{\epsilon},
\qquad \boldsymbol{\epsilon}\sim\mathcal{N}(\boldsymbol{0},\boldsymbol{I}).
\end{aligned}
\label{eq:method_initialization}
\end{equation}
The standard deviation of the added noise is thus scaled with the starting time as $\tau_{\omega}\gamma_{\omega}$, where $\gamma_{\omega}$ is unchanged from training. For the same noise realization, the difference between this initialization and the corresponding training state is $(1-\tau_{\omega})(\boldsymbol{c}_{\omega}-\boldsymbol{x}^{\star})$. This construction preserves the scale of the added noise at the selected time, while the state discrepancy is determined by the initial reconstruction error.

The three tasks use the following time selection rules.

\noindent\textbf{1) Low-Dose CT:}
Low-dose degradation primarily manifests as increased noise. We therefore estimate the noise level from the initial reconstruction as $\boldsymbol{q}_{\mathrm{LD}}=\widehat{\sigma}=\mathcal{E}(\boldsymbol{c}_{\mathrm{LD}})$. Specifically, the image is divided into local $2\times2$ pixel blocks, and differences between sorted pixel values within each block are computed. These differences are calibrated to the Gaussian noise scale and aggregated across blocks to estimate the overall noise standard deviation. The starting time is set to
\begin{equation}
\tau_{\mathrm{LD}}
=\operatorname{clip}\!\left(
\frac{\widehat{\sigma}}{\sigma_{\mathrm{ref}}},0,1\right),
\label{eq:method_ld_start}
\end{equation}
where $\sigma_{\mathrm{ref}}>0$ is a reference noise level defined under the same image normalization, and $\operatorname{clip}(\cdot,0,1)$ restricts the value to $[0,1]$. A higher estimated noise level yields a longer restoration interval and adjusts the initial perturbation accordingly through \eqref{eq:method_initialization}.

\noindent\textbf{2) Sparse-View CT:}
Sparse-view degradation arises from angular undersampling over the full scan range. Let $\Theta$ denote the reference angle set for a full scan and $\Omega\subset\Theta$ the acquired angle set. The missing-view fraction $\boldsymbol{q}_{\mathrm{SV}}=\eta$ determines the starting time:
\begin{equation}
\tau_{\mathrm{SV}}=\eta
=1-\frac{|\Omega|}{|\Theta|}.
\label{eq:method_sv_start}
\end{equation}
For a fixed scan range and sampling scheme, fewer views yield a larger starting time; with more views available, restoration starts closer to the target endpoint.

\noindent\textbf{3) Limited-Angle CT:}
Limited-angle degradation arises from a missing contiguous angular range. The acquisition interval is specified by its starting and ending angles, $\phi_{\mathrm{s}}$ and $\phi_{\mathrm{e}}$, with the descriptor
\begin{equation}
\boldsymbol{q}_{\mathrm{LA}}
=\bigl[\sin\phi_{\mathrm{s}},\cos\phi_{\mathrm{s}},
\sin\phi_{\mathrm{e}},\cos\phi_{\mathrm{e}}\bigr]^{\mathsf T}.
\label{eq:method_la_descriptor}
\end{equation}
The actual angular coverage $\Delta\phi$ is computed according to the known scan direction and normalized by the reference full-scan range $\Delta\phi_{\mathrm{ref}}$ under the same imaging geometry:
\begin{equation}
\tau_{\mathrm{LA}}
=1-\frac{\Delta\phi}{\Delta\phi_{\mathrm{ref}}}.
\label{eq:method_la_start}
\end{equation}
This rule controls the restoration interval through the missing-angle fraction: smaller angular coverage yields a larger starting time.

For all three tasks, the selected time is mapped to the nearest positive node in $\mathcal{S}$. The discretized time, still denoted by $\tau_{\omega}$, is used for initialization in \eqref{eq:method_initialization}. We then take the decreasing nodes from $\tau_{\omega}$ to $0$ on the predefined grid, denoted by $\tau_{\omega}=t_0>t_1>\cdots>t_K=0$, set $\boldsymbol{x}_{t_0}=\boldsymbol{x}_{\mathrm{init}}$, and perform Euler updates using the task-specific student network:
\begin{equation}
\boldsymbol{x}_{t_{k+1}}
=\boldsymbol{x}_{t_k}
+(t_{k+1}-t_k)\,
\boldsymbol{v}_{\theta_{\omega}}
(\boldsymbol{x}_{t_k},t_k\mid\boldsymbol{c}_{\omega}),
\label{eq:method_euler_inference}
\end{equation}
where $k=0,\ldots,K-1$. This update uses the same time step and discretization as feedback training. The random perturbation is sampled only once, at initialization, and the final reconstruction is $\widehat{\boldsymbol{x}}_{\omega}=\boldsymbol{x}_{t_K}$. Algorithm~\ref{alg:FFAI_inference} summarizes the inference procedure.

\begin{algorithm}[t]
\caption{Degradation-Adaptive Inference}
\label{alg:FFAI_inference}
\begin{algorithmic}[1]
\REQUIRE Task $\omega$, test projections $\boldsymbol{y}_{\omega}$ and acquisition information, student parameters $\theta_{\omega}$, inference grid $\mathcal{S}$
\ENSURE Reconstructed image $\widehat{\boldsymbol{x}}_{\omega}$
\STATE Compute the initial reconstruction $\boldsymbol{c}_{\omega}=\mathcal{B}_{\omega}(\boldsymbol{y}_{\omega})$
\STATE Obtain the task-specific degradation descriptor $\boldsymbol{q}_{\omega}$
\STATE Select $\tau_{\omega}$ using the task-specific time rule and map it to the nearest positive inference node
\STATE Sample $\boldsymbol{\epsilon}\sim\mathcal{N}(\boldsymbol{0},\boldsymbol{I})$
\STATE Construct $\boldsymbol{x}_{\mathrm{init}}$ using \eqref{eq:method_initialization}
\STATE Select decreasing nodes $\tau_{\omega}=t_0>\cdots>t_K=0$ and set $\boldsymbol{x}_{t_0}=\boldsymbol{x}_{\mathrm{init}}$
\FOR{$k=0$ to $K-1$}
    \STATE Update to $\boldsymbol{x}_{t_{k+1}}$ using \eqref{eq:method_euler_inference}
\ENDFOR
\STATE $\widehat{\boldsymbol{x}}_{\omega}\leftarrow\boldsymbol{x}_{t_K}$
\end{algorithmic}
\end{algorithm}

\begin{table*}[t]
    \centering
    \caption{Quantitative results for quarter-dose CT reconstruction (mean $\pm$ standard deviation).
    SSIM is reported as a percentage. Bold and underlined values indicate the best and second-best means, respectively.}
    \label{tab:ldct_with_tamp}
    \begingroup
    \small
    \setlength{\tabcolsep}{4pt}
    \renewcommand{\arraystretch}{1.2}
    \resizebox{\textwidth}{!}{%
    \begin{tabular}{@{}lccccccc@{}}
        \toprule
        Metric
        & FBP
        & Noise2Noise
        & MTD-GAN
        & FoundDiff
        & TAMP
        & FORCE
        & \textbf{FFAI} \\
        \midrule
        PSNR (dB) $\uparrow$
        & 40.99\,$\pm$\,0.08
        & 41.79\,$\pm$\,0.23
        & \underline{46.03\,$\pm$\,0.08}
        & 45.51\,$\pm$\,0.06
        & 42.45\,$\pm$\,1.13
        & 40.81\,$\pm$\,0.33
        & \textbf{48.30\,$\pm$\,0.05} \\
        SSIM (\%) $\uparrow$
        & 93.70\,$\pm$\,0.14
        & 95.50\,$\pm$\,0.10
        & 97.81\,$\pm$\,0.05
        & \underline{97.87\,$\pm$\,0.53}
        & 96.43\,$\pm$\,1.39
        & 96.04\,$\pm$\,0.47
        & \textbf{97.96\,$\pm$\,0.39} \\
        \bottomrule
    \end{tabular}%
    }
    \endgroup
\end{table*}

\begin{figure*}[!h]
    \centering
    \includegraphics[width=\textwidth]{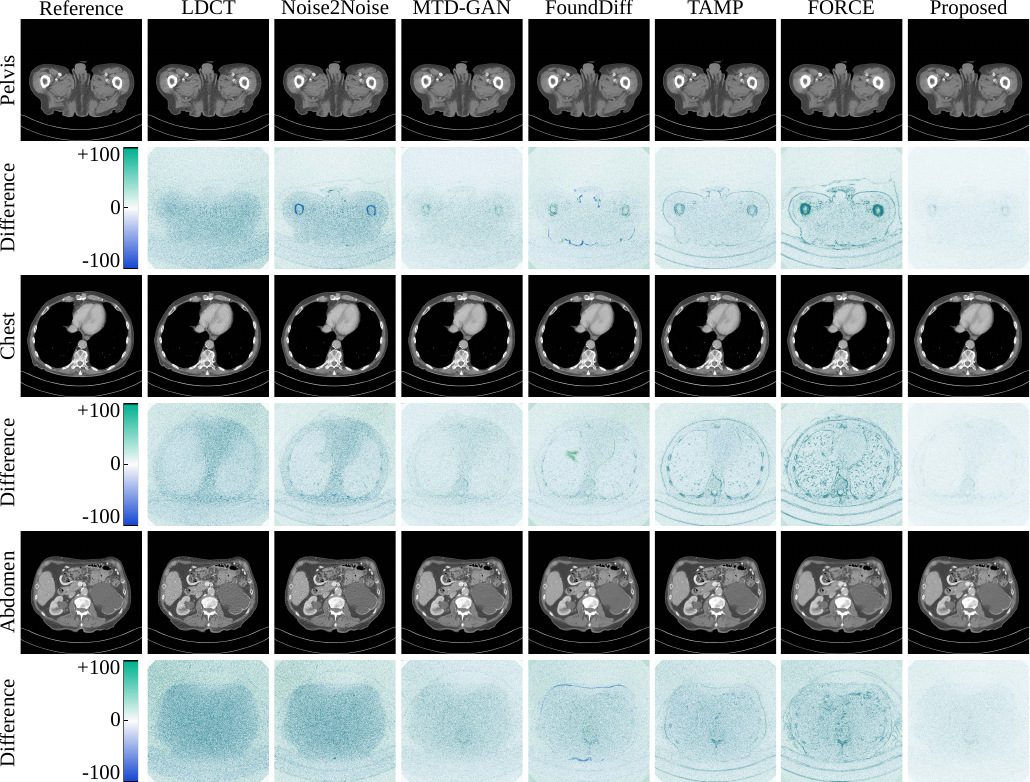}
    \caption{Quarter-dose CT reconstructions obtained with different methods.
    Representative pelvic, thoracic, and abdominal slices are shown from top to bottom.
    In each group, the upper row shows CT images, and the lower row shows difference maps relative to the full-dose reference.
    Reference denotes the full-dose reference image, LDCT denotes the quarter-dose input image,
    and Proposed denotes FFAI.
    The display window is $[-260,340]$~HU,
    and the color scale for the difference maps spans $[-100,100]$.}
    \label{fig:ldct_comparison}
\end{figure*}

\begin{table*}[t]
    \centering
    \caption{Quantitative results for $60^\circ$ and $90^\circ$ limited-angle CT
    reconstruction (mean $\pm$ standard deviation).
    SSIM is reported as a percentage. Bold and underlined values indicate
    the best and second-best means, respectively, for each angular range.}
    \label{tab:limited_angle}

    \begingroup
    \small
    \setlength{\tabcolsep}{12pt}
    \renewcommand{\arraystretch}{1.2}

    \begin{tabular}{@{}lcccc@{}}
        \toprule
        \multirow{2}{*}{Method}
        & \multicolumn{2}{c}{$60^\circ$}
        & \multicolumn{2}{c}{$90^\circ$} \\
        \cmidrule(lr){2-3}
        \cmidrule(lr){4-5}
        & PSNR (dB) $\uparrow$
        & SSIM (\%) $\uparrow$
        & PSNR (dB) $\uparrow$
        & SSIM (\%) $\uparrow$ \\
        \midrule

        FBP
        & 17.28\,$\pm$\,1.97
        & 53.55\,$\pm$\,3.06
        & 19.62\,$\pm$\,1.89
        & 64.09\,$\pm$\,4.05 \\

        IRON
        & 31.57\,$\pm$\,1.70
        & 90.36\,$\pm$\,3.26
        & 35.61\,$\pm$\,3.65
        & 94.60\,$\pm$\,2.66 \\

        ProCT
        & \underline{34.98\,$\pm$\,1.59}
        & \underline{92.56\,$\pm$\,2.00}
        & \underline{38.83\,$\pm$\,1.90}
        & \underline{95.67\,$\pm$\,1.52} \\

        TLIR
        & 20.33\,$\pm$\,1.14
        & 54.57\,$\pm$\,12.04
        & 25.23\,$\pm$\,3.02
        & 46.09\,$\pm$\,3.59 \\

        TAMP
        & 20.79\,$\pm$\,1.75
        & 60.87\,$\pm$\,3.33
        & 24.57\,$\pm$\,3.01
        & 77.62\,$\pm$\,2.53 \\

        PWD
        & 28.87\,$\pm$\,2.39
        & 74.09\,$\pm$\,4.89
        & 31.55\,$\pm$\,2.77
        & 82.84\,$\pm$\,8.26 \\

        \midrule
        \textbf{FFAI}
        & \textbf{38.94\,$\pm$\,1.79}
        & \textbf{94.28\,$\pm$\,2.07}
        & \textbf{40.92\,$\pm$\,1.89}
        & \textbf{95.97\,$\pm$\,1.57} \\

        \bottomrule
    \end{tabular}
    \endgroup
\end{table*}

\begin{figure*}[!t]
    \centering
    \includegraphics[width=\textwidth]{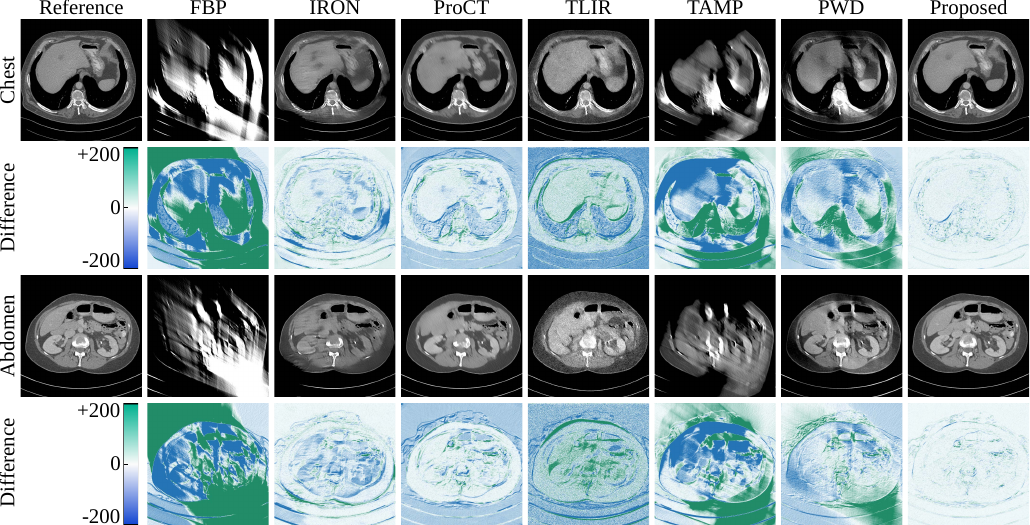}
    \caption{Reconstructions obtained with different methods under the $60^\circ$ limited-angle setting.
    The upper and lower groups show representative thoracic and abdominal slices, respectively.
    In each group, the upper row shows CT images, and the lower row shows difference maps relative to the reference.
    Reference denotes the full-dose reference image, and Proposed denotes FFAI.
    The display window is $[-260,340]$~HU,
    and the color scale for the difference maps spans $[-200,200]$.}
    \label{fig:lact60_comparison}
\end{figure*}

\begin{figure*}[!t]
    \centering
    \includegraphics[width=\textwidth]{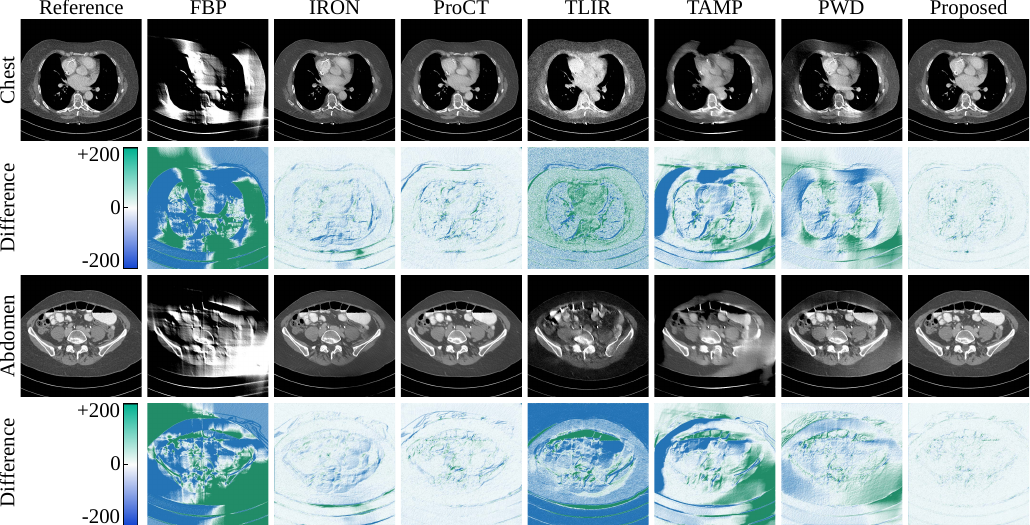}
    \caption{Reconstructions obtained with different methods under the $90^\circ$ limited-angle setting.
    The upper and lower groups show representative thoracic and abdominal slices, respectively.
    In each group, the upper row shows CT images, and the lower row shows difference maps relative to the reference.
    Reference denotes the full-dose reference image, and Proposed denotes FFAI.
    The display window is $[-260,340]$~HU,
    and the color scale for the difference maps spans $[-200,200]$.}
    \label{fig:lact90_comparison}
\end{figure*}

\section{Experiments and Results}

We evaluated FFAI on the Mayo 2016 low-dose CT dataset for low-dose, limited-angle, and sparse-view CT reconstruction. All two-dimensional slices had a matrix size of $512\times512$ pixels. The low-dose experiments used the official paired quarter-dose (QD25) and full-dose (FD) images as inputs and references, respectively. The training set comprised $1923$ slice pairs from $8$ patients (L067, L096, L109, L143, L192, L286, L291, and L310), and the test set included all $211$ slice pairs from patient L506. For limited-angle CT, $1156$ slices from $5$ patients (L067, L096, L109, L143, and L192) were used for training. The test set comprised $210$ slices from patient L286, of which $50$ were randomly selected for evaluation. Two angular ranges, $60^\circ$ and $90^\circ$, were considered by retaining contiguous projections within $[0^\circ,60^\circ)$ and $[0^\circ,90^\circ)$, respectively. The sparse-view experiments used the same $5$ training patients as the limited-angle experiments and $525$ slices from patient L286 for evaluation. Two sparse-view settings were generated by uniformly selecting $40$ or $80$ projection views over the full $360^\circ$ range. The corresponding full-dose CT images served as references in both the limited-angle and sparse-view experiments.

\subsection{Low-Dose CT Study}
\label{subsec:ldct_study}

The low-dose experiments used the official quarter-dose images from the Mayo 2016 dataset, with the corresponding full-dose images as references. FFAI was compared with Noise2Noise~\cite{lehtinen2018noise2noise}, MTD-GAN~\cite{kyung2025robustgan}, FoundDiff~\cite{chen2026founddiff}, TAMP~\cite{ge2026tamp}, and FORCE~\cite{xia2026force}, with FBP included as a baseline. Table~\ref{tab:ldct_with_tamp} reports the mean and standard deviation of PSNR and SSIM for each method. FFAI achieved a mean PSNR of $48.30$~dB and a mean SSIM of $97.96\%$, the highest values in this comparison. FFAI exceeded MTD-GAN, which ranked second in PSNR, by $2.27$~dB and FoundDiff, which ranked second in SSIM, by $0.09$ percentage points.

Fig.~\ref{fig:ldct_comparison} shows representative pelvic, thoracic, and abdominal slices and their difference maps relative to the full-dose references. Noticeable granular residuals remained after Noise2Noise processing. Tissue contours were visible to varying degrees in the difference maps of FoundDiff, TAMP, and FORCE, indicating residual structure-related differences from the reference images. MTD-GAN and FFAI produced smaller differences overall, with FFAI more closely matching the references in the displayed regions. These results indicate that FFAI effectively suppresses noise while preserving anatomical structures under the evaluated low-dose setting.

\subsection{Limited-Angle CT Study}
\label{subsec:lact_study}

The limited-angle experiments considered two contiguous projection ranges of $60^\circ$ and $90^\circ$, using the corresponding full-dose CT images as references. FFAI was compared with IRON~\cite{pan2024iron}, ProCT~\cite{ma2026proct}, TLIR~\cite{li2025tlir}, TAMP~\cite{ge2026tamp}, and PWD~\cite{liu2026pwd}, with FBP included as a baseline. Table~\ref{tab:limited_angle} summarizes the quantitative results for both angular ranges. FFAI achieved mean PSNR and SSIM values of $38.94$~dB and $94.28\%$ at $60^\circ$, and $40.92$~dB and $95.97\%$ at $90^\circ$, respectively, attaining the highest means for both metrics in both settings. Compared with the second-best method, ProCT, FFAI improved PSNR by $3.96$~dB and $2.09$~dB and SSIM by $1.72$ and $0.30$ percentage points, respectively.

Figs.~\ref{fig:lact60_comparison} and~\ref{fig:lact90_comparison} show representative thoracic and abdominal reconstructions and their difference maps. FBP images exhibited pronounced directional streaks and intensity distortions. Strong artifacts remained in some regions of the TAMP and PWD reconstructions, and the TLIR difference maps also showed pronounced structure-related residuals. IRON and ProCT improved the recovery of tissue contours, but distinct tissue boundaries remained visible in their difference maps. FFAI better preserved tissue morphology in the displayed slices and produced weaker structure-related residuals. Together with the quantitative results, these examples indicate that FFAI provides a favorable balance between artifact suppression and structural recovery under both limited-angle settings.

\subsection{Sparse-View CT Study}
\label{subsec:svct_study}

\begin{table*}[t]
    \centering
    \caption{Quantitative results for sparse-view CT reconstruction with $40$ and $80$ views (mean $\pm$ standard deviation).
    SSIM is reported as a percentage. Bold and underlined values indicate the best and second-best means, respectively, for each view setting.}
    \label{tab:sparse_view}
    \begingroup
    \small
    \setlength{\tabcolsep}{12pt}
    \renewcommand{\arraystretch}{1.2}
    \begin{tabular}{@{}lcccc@{}}
        \toprule
        \multirow{2}{*}{Method}
        & \multicolumn{2}{c}{40 views}
        & \multicolumn{2}{c}{80 views} \\
        \cmidrule(lr){2-3}
        \cmidrule(lr){4-5}
        & PSNR (dB) $\uparrow$
        & SSIM (\%) $\uparrow$
        & PSNR (dB) $\uparrow$
        & SSIM (\%) $\uparrow$ \\
        \midrule
        FBP
        & 26.65\,$\pm$\,1.07
        & 48.52\,$\pm$\,2.82
        & 32.25\,$\pm$\,0.79
        & 73.07\,$\pm$\,1.62 \\
        LipCT
        & 37.29\,$\pm$\,0.56
        & 87.27\,$\pm$\,0.97
        & 39.25\,$\pm$\,0.21
        & 90.72\,$\pm$\,0.86 \\
        MetaInv
        & 35.56\,$\pm$\,1.76
        & 84.50\,$\pm$\,2.34
        & 37.08\,$\pm$\,1.14
        & 88.08\,$\pm$\,0.97 \\
        DPS
        & 32.33\,$\pm$\,0.40
        & 74.75\,$\pm$\,1.14
        & 32.42\,$\pm$\,0.43
        & 75.04\,$\pm$\,0.96 \\
        TAMP
        & 30.25\,$\pm$\,0.65
        & 80.55\,$\pm$\,1.07
        & 36.00\,$\pm$\,0.49
        & 89.11\,$\pm$\,0.95 \\
        FORCE
        & \underline{37.55\,$\pm$\,0.64}
        & \underline{88.76\,$\pm$\,0.94}
        & \underline{39.86\,$\pm$\,0.35}
        & \underline{92.77\,$\pm$\,0.78} \\
        \midrule
        \textbf{FFAI}
        & \textbf{39.08\,$\pm$\,0.31}
        & \textbf{90.55\,$\pm$\,1.18}
        & \textbf{41.67\,$\pm$\,0.36}
        & \textbf{94.68\,$\pm$\,1.06} \\
        \bottomrule
    \end{tabular}
    \endgroup
\end{table*}

\begin{figure*}[!t]
    \centering
    \includegraphics[width=\textwidth]{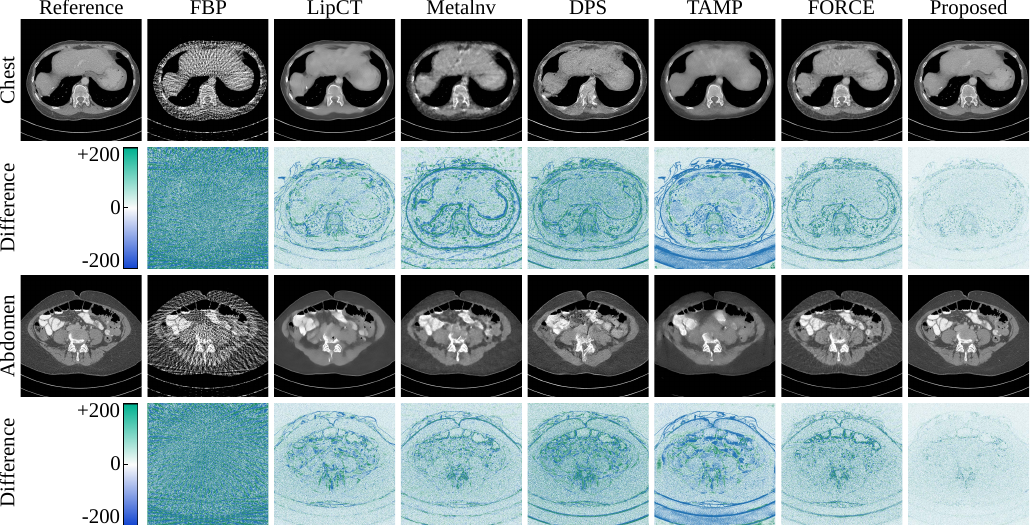}
    \caption{CT reconstructions obtained with different methods under the $40$-view sparse-sampling setting.
    The upper and lower groups show representative thoracic and abdominal slices, respectively.
    In each group, the upper row shows CT images, and the lower row shows difference maps relative to the reference.
    Reference denotes the full-dose reference image, and Proposed denotes FFAI.
    The display window is $[-260,340]$~HU,
    and the color scale for the difference maps spans $[-200,200]$.}
    \label{fig:svct40_comparison}
\end{figure*}

\begin{figure*}[!t]
    \centering
    \includegraphics[width=\textwidth]{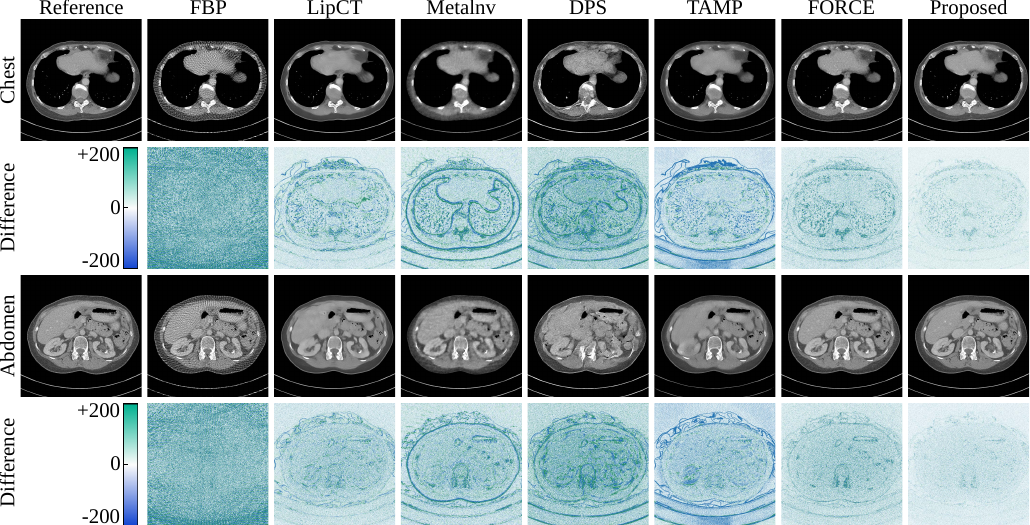}
    \caption{CT reconstructions obtained with different methods under the $80$-view sparse-sampling setting.
    The upper and lower groups show representative thoracic and abdominal slices, respectively.
    In each group, the upper row shows CT images, and the lower row shows difference maps relative to the reference.
    Reference denotes the full-dose reference image, and Proposed denotes FFAI.
    The display window is $[-260,340]$~HU,
    and the color scale for the difference maps spans $[-200,200]$.}
    \label{fig:svct80_comparison}
\end{figure*}

The sparse-view experiments used $40$ and $80$ projection views uniformly sampled over the full $360^\circ$ range, with the corresponding full-dose CT images as references. FFAI was compared with LipCT~\cite{shi2026promptct}, MetaInv~\cite{zhang2021metainv}, DPS~\cite{chung2023dps}, TAMP~\cite{ge2026tamp}, and FORCE~\cite{xia2026force}, with FBP included as a baseline. Table~\ref{tab:sparse_view} summarizes the quantitative results for both view settings. FFAI achieved mean PSNR and SSIM values of $39.08$~dB and $90.55\%$ with $40$ views, and $41.67$~dB and $94.68\%$ with $80$ views, respectively, attaining the highest means for both metrics. Compared with the second-best method, FORCE, FFAI improved PSNR by $1.53$~dB and $1.81$~dB and SSIM by $1.79$ and $1.91$ percentage points, respectively.

Figs.~\ref{fig:svct40_comparison} and~\ref{fig:svct80_comparison} show representative thoracic and abdominal reconstructions and their difference maps. Pronounced streak artifacts were present in the FBP images. LipCT and TAMP suppressed many streaks, but some tissue boundaries and internal details remained blurred. Local structural distortions or granular residuals were still visible in the MetaInv and DPS reconstructions. FORCE recovered more tissue details, although tissue contours remained visible in its difference maps. FFAI better preserved tissue boundaries and internal structures in the displayed slices and produced weaker structure-related residuals in the difference maps. Together with the quantitative results, these examples indicate that FFAI provides a favorable balance between streak suppression and structural recovery under both sparse-view settings.

\subsection{Ablation Study}
\label{subsec:ablation_study}

\begin{figure}[!t]
    \centering
    \includegraphics[width=0.9\columnwidth]{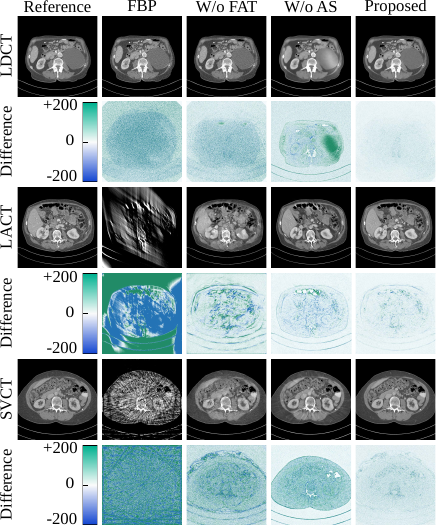}
    \caption{CT reconstructions obtained with different ablation variants.
    From top to bottom, the three groups correspond to quarter-dose, $60^\circ$ limited-angle, and $40$-view sparse-view experiments.
    In each group, the upper row shows the reconstructions and the lower row shows their differences from the reference images.
    W/o FAT and W/o AS denote the variants without feedback-aware training and with a fixed starting point, respectively.
    Proposed denotes the full FFAI model.
    The color scale for the difference maps is $[-200,200]$.}
    \label{fig:Ablation}
\end{figure}

\begin{table*}[t]
    \centering
    \caption{Ablation results for quarter-dose, $60^\circ$ limited-angle, and $40$-view sparse-view CT
    reconstruction (mean $\pm$ standard deviation).
    FAT and AS denote feedback-aware training and an adaptive starting point, respectively.
    $\checkmark$ and $\times$ indicate whether each component is enabled or removed;
    removing AS corresponds to using a fixed starting point.
    SSIM is reported as a percentage, and boldface indicates the best mean values.}
    \label{tab:ablation}
    \begingroup
    \footnotesize
    \setlength{\tabcolsep}{5.2pt}
    \renewcommand{\arraystretch}{1.15}
    \resizebox{\textwidth}{!}{%
    \begin{tabular}{@{}lcccccccc@{}}
        \toprule
        \multirow{2}{*}{Variant}
        & \multirow{2}{*}{FAT}
        & \multirow{2}{*}{AS}
        & \multicolumn{2}{c}{LDCT (25\%)}
        & \multicolumn{2}{c}{LACT ($60^\circ$)}
        & \multicolumn{2}{c}{SVCT (40 views)} \\
        \cmidrule(lr){4-5}
        \cmidrule(lr){6-7}
        \cmidrule(lr){8-9}
        & & &
        PSNR (dB) $\uparrow$ & SSIM (\%) $\uparrow$ &
        PSNR (dB) $\uparrow$ & SSIM (\%) $\uparrow$ &
        PSNR (dB) $\uparrow$ & SSIM (\%) $\uparrow$ \\
        \midrule

        W/o FAT
        & $\times$ & $\checkmark$
        & 41.44\,$\pm$\,1.60
        & 93.91\,$\pm$\,2.17
        & 28.97\,$\pm$\,1.80
        & 74.35\,$\pm$\,2.98
        & 34.87\,$\pm$\,0.68
        & 79.88\,$\pm$\,1.20 \\

        W/o AS
        & $\checkmark$ & $\times$
        & 30.71\,$\pm$\,1.77
        & 81.85\,$\pm$\,2.20
        & 32.69\,$\pm$\,0.94
        & 82.44\,$\pm$\,1.39
        & 35.42\,$\pm$\,0.63
        & 81.12\,$\pm$\,1.01 \\

        \midrule
        \textbf{FFAI}
        & $\checkmark$ & $\checkmark$
        & \textbf{48.30\,$\pm$\,0.05}
        & \textbf{97.96\,$\pm$\,0.39}
        & \textbf{38.94\,$\pm$\,1.79}
        & \textbf{94.28\,$\pm$\,2.07}
        & \textbf{39.08\,$\pm$\,0.31}
        & \textbf{90.55\,$\pm$\,1.18} \\
        \bottomrule
    \end{tabular}%
    }
    \endgroup
\end{table*}

To assess the contributions of error correction and adaptive initialization, we evaluated two ablation variants: removing feedback-aware training (W/o FAT) and replacing the adaptive starting point with a fixed one (W/o AS). FAT denotes the feedback-aware mechanism used during training, whereas AS selects the inference starting point based on degradation severity. Experiments were conducted under quarter-dose, $60^\circ$ limited-angle, and $40$-view sparse-view settings. Table~\ref{tab:ablation} presents the quantitative results.

Removing FAT reduced the mean PSNR by $6.86$, $9.97$, and $4.21$~dB for the low-dose, limited-angle, and sparse-view tasks, respectively, indicating that feedback-aware training improves reconstruction quality across all three tasks. Replacing the adaptive starting point with a fixed one reduced the mean PSNR by $17.59$, $6.25$, and $3.66$~dB, respectively, showing that degradation-adaptive initialization outperforms the fixed starting configuration used here. The full model achieved the highest mean PSNR and SSIM in all three tasks.

Fig.~\ref{fig:Ablation} shows the reconstructions and difference maps for each variant. Without FAT, appreciable noise or structure-related residuals remained. With a fixed starting point, tissue contours remained visible in the difference maps. In comparison, the full model produced smaller differences overall and better preserved tissue boundaries and internal structures in the displayed slices.

\subsection{Generalizability Study}

To assess cross-dataset generalizability, we evaluated the proposed method on the CT-ORG \cite{rister2020ctorg} and LDCT-and-Projection-data \cite{moen2021lowdose} datasets provided by TCIA. The sparse-view and limited-angle experiments used CT-ORG, which includes abdominal and whole-body scans acquired under varying conditions, including contrast-enhanced and noncontrast examinations. We selected one slice from each of 140 CT scans to form a shared reference image set for sparse-view reconstruction with 80 projection views and limited-angle reconstruction over a $90^\circ$ angular range. The low-dose experiments used 20 abdominal cases from LDCT-and-Projection-data. Five image pairs were selected per case by stratified sampling along the slice axis, yielding 100 pairs in total. The inputs were the dataset-provided reconstructions simulated at $25\%$ of routine dose by projection-domain noise insertion, and the references were the corresponding routine-dose images. All images were standardized to $512 \times 512$ pixels, and reconstruction quality was evaluated using PSNR and SSIM.

\begin{figure}[!t]
    \centering
    \includegraphics[width=0.9\columnwidth]{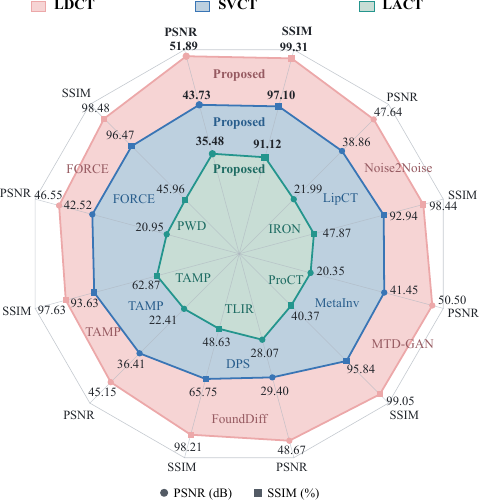}
    \caption{Quantitative comparison of cross-dataset generalizability. Low-dose CT (LDCT) uses $25\%$ of routine dose, sparse-view CT (SVCT) uses 80 projection views, and limited-angle CT (LACT) uses a $90^\circ$ angular range. Circles and squares indicate PSNR (dB) and SSIM ($\%$), respectively. Proposed denotes the proposed method.}
    \label{fig:Generalization}
\end{figure}

As shown in Fig.~\ref{fig:Generalization}, the proposed method achieved the highest PSNR and SSIM in all three tasks. For low-dose reconstruction, it achieved 51.89~dB and $99.31\%$, exceeding MTD-GAN, the best-performing comparison method, by 1.39~dB and 0.26 percentage points, respectively. For sparse-view reconstruction, it achieved 43.73~dB and $97.10\%$, exceeding FORCE by 1.21~dB and 0.63 percentage points, respectively.

The advantage was more pronounced for limited-angle reconstruction. The proposed method achieved a PSNR of 35.48~dB and an SSIM of $91.12\%$. Its PSNR exceeded that of the second-best method, TLIR, by 7.41~dB, while its SSIM exceeded that of the second-best method, TAMP, by 28.25 percentage points. These results show that the proposed method retained its reconstruction advantage on the additional test datasets, supporting its generalizability across the three CT reconstruction tasks.

\section{Discussion and Conclusion}
\label{sec:discussion}

We proposed FFAI, a conditional flow matching method for low-dose, sparse-view, and limited-angle CT reconstruction. Conditioned on an initial degraded reconstruction, FFAI incorporates teacher-generated feedback states during training to learn to recover high-quality images from states shifted by prediction errors. During inference, it selects the starting time based on input degradation severity and constructs the corresponding initial state. By combining feedback-aware training with degradation-adaptive initialization, FFAI provides a common modeling and reconstruction procedure for all three tasks.

Experiments on public benchmark datasets showed that FFAI outperformed the comparison methods quantitatively across reconstruction tasks while providing good visual quality. It effectively suppressed noise and artifacts while preserving tissue structures and details, demonstrating its ability to handle different types of degradation.

Despite these results, FFAI has several limitations. It currently requires task-specific settings for some inference parameters and a separately trained model for each task. The present study also focuses primarily on two-dimensional reconstruction, and applicability to a broader range of imaging conditions requires further validation. Future work may address these limitations in the following directions:
\begin{enumerate}
    \item \textbf{Automated inference configuration.}
    Acquisition information and image features could guide the joint selection of the starting time, noise scale, and number of inference steps. This may reduce manual task-specific configuration and enable more flexible tradeoffs between reconstruction quality and computational cost.

    \item \textbf{Cross-task parameter sharing.}
    A common representation of degradation conditions could be incorporated into the framework to support joint training and parameter sharing across tasks. Exploiting commonalities among reconstruction tasks may reduce the cost of training and maintaining separate models and facilitate adaptation to combined degradations.

    \item \textbf{Extension to three-dimensional reconstruction.}
    Extending the method to three-dimensional reconstruction could exploit structural information from adjacent slices to improve spatial continuity in volumetric images. Efficient network architectures and inference strategies could also help control the GPU memory requirements and computational costs of volumetric reconstruction.

    \item \textbf{Adaptation to different imaging conditions.}
    Data from more scanners, acquisition protocols, and anatomical regions could support further investigation of adaptability to different imaging conditions. Multicenter validation, reader studies, and downstream analysis tasks could provide a more comprehensive assessment of reconstruction quality and practical utility.
\end{enumerate}

\bibliographystyle{IEEEtran}
\bibliography{ref}

\end{document}